%% file: main.tex
\def\year{2021}\relax
\documentclass[letterpaper]{article} 
\usepackage{aaai21}  
\usepackage{times}  
\usepackage{helvet} 
\usepackage{courier}  
\usepackage[hyphens]{url}  
\usepackage{graphicx} 
\def\UrlFont{\rm}  
\usepackage{natbib}  
\usepackage{caption} 
\usepackage{amsfonts}
\usepackage{multirow}
\usepackage{amsmath}
\usepackage{times}
\usepackage{soul}
\usepackage{url}
\usepackage{amsthm}
\usepackage{booktabs}
\usepackage{algorithm}
\usepackage{algorithmic}
\usepackage{float}
\usepackage{amssymb}
\title{Multi-view Inference for Relation Extraction with Uncertain Knowledge}
\author{Bo Li$^{1,2}$, Wei Ye$^{1}$\thanks{{} {} Corresponding author.}, Canming Huang$^{3}$, Shikun Zhang$^{1}$ \\}
\affiliations{
$^1$National Engineering Research Center 
  for Software Engineering, Peking University \\
  $^2$School of Software and Microelectronics, Peking University\\
  $^3$Beijing University of Posts and Telecommunications\\
  {deepblue.lb@stu.pku.edu.cn}, { wye@pku.edu.cn}, \\{huangzzk@bupt.edu.cn}, {zhangsk@pku.edu.cn}
}
\begin{document}

\maketitle

\begin{abstract}

Knowledge graphs (KGs) are widely used to facilitate relation extraction (RE) tasks. While most previous RE methods focus on leveraging deterministic KGs, uncertain KGs, which assign a confidence score for each relation instance, can provide prior probability distributions of relational facts as valuable external knowledge for RE models. This paper proposes to exploit uncertain knowledge to improve relation extraction. Specifically, we introduce ProBase, an uncertain KG that indicates to what extent a target entity belongs to a concept, into our RE architecture. We then design a novel multi-view inference framework to systematically integrate local context and global knowledge across three views: mention-, entity- and concept-view. The experimental results show that our model achieves competitive performances on both sentence- and document-level relation extraction, which verifies the effectiveness of introducing uncertain knowledge and the multi-view inference framework that we design.

\end{abstract}
\input{1-introduction-12.14}

\input{3-input}

\input{4-methods}
\input{5-experiments}
\input{2-related-work}
\section{Conclusion and Future Work}

We have presented MIUK, a \textbf{M}ulti-view \textbf{I}nference framework for relation extraction with \textbf{U}ncertain \textbf{K}nowledge. MIUK introduces ProBase, an uncertain KG, into relation extraction pipeline for the first time. To effectively incorporate ProBase, we have designed a multi-view inference mechanism that integrates local context and global knowledge across mention-,  entity-,  and concept-view. Results of extensive experiments on both sentence- and document-level relation extraction tasks can verify the effectiveness of our method. We have also built a corpus with high-quality descriptions of entities and concepts, serving as a valuable external knowledge resource for relation extraction and many other NLP tasks. We believe it could be a future research direction of relation extraction to further investigate the interactions among mentions, entities, and concepts.


\section*{Acknowledgements}
This research is supported by the National Key Research And Development Program of China (No. 2019YFB1405802). We would also like to thank Handan Institute of Innovation, Peking University for their support of our work.

\bibliography{ref}

\end{document}

%% file: 1-introduction-12.14.tex
\section{Introduction}


The goal of relation extraction (RE) is to classify the semantic relation between entities in a given context. It has plenty of practical applications, such as question answering \cite{DBLP:conf/acl/YuYHSXZ17} and information retrieval \cite{DBLP:conf/sigir/KadryD17}. Knowledge graphs (KGs) \cite{DBLP:conf/sigmod/BollackerEPST08,ruppenhofer2006framenet,DBLP:conf/aaai/SpeerCH17,DBLP:conf/sigmod/WuLWZ12}, which contain pre-defined nodes (usually entities or concepts) and their relations, have been widely incorporated into RE tasks to integrate various prior knowledge in recent years.

The interaction between KGs and RE lies in two main aspects. On the one hand, we can use relational facts in exiting KGs to assign relation labels for entity pairs in unlabeled corpora to build datasets for distant supervision RE \cite{DBLP:conf/acl/MintzBSJ09}. On the other hand, we can leverage the external knowledge in KGs to boost the performance of RE models, which is the case of this paper. Generally, there are two ways to integrate prior knowledge into RE. One way is to utilize structured information explicitly. For example, \citet{DBLP:conf/naacl/CanLHC19} retrieved the synonyms of each entity from WordNet\footnote{https://wordnet.princeton.edu/}; \citet{DBLP:conf/coling/LeiCLDY0S18} extracted the n-gram text matching words of each entity from FreeBase \cite{DBLP:conf/sigmod/BollackerEPST08}; \citet{DBLP:conf/emnlp/LiMYL19} used relative semantic frames from FrameNet \cite{ruppenhofer2006framenet}. The other way is to explore the latent semantics of KGs. Researchers may incorporate pre-trained entity and relation embeddings \cite{DBLP:conf/emnlp/WangZWZCZZC18, DBLP:conf/emnlp/HuZSNGY19},  e.g., from TransE \cite{DBLP:conf/nips/BordesUGWY13}, or jointly learn the entity embedding and RE models' parameters \cite{DBLP:conf/aaai/Han0S18,DBLP:conf/naacl/XuB19}.

Despite various ways of leveraging KGs, most previous RE methods focus on deterministic KGs, where a specific relation either exists between two entities or not. However, in real-world scenarios, prior knowledge resources may contain inconsistent, ambiguous, and uncertain information \cite{DBLP:conf/sigmod/WuLWZ12}.  It would not be easy to capture highly-related external information if we treat all relations between entities as equally important. Another type of KGs, called Uncertain KGs \cite{DBLP:conf/aaai/ChenCSSZ19} such as ConceptNet \cite{DBLP:conf/aaai/SpeerCH17} and ProBase \cite{DBLP:conf/sigmod/WuLWZ12}, come to the rescue. Given two words (typically two entities or an entity and a concept) and a relation, uncertain KGs provide a confidence score that measures the possibility of the two words having the given relation. This inspires us to exploit the prior probability distribution of relational facts in uncertain KGs to improve relation extraction.

As a representative uncertain KG, ProBase provides \textit{IsA} relation between an entity and a concept, indicating to what extent the target entity belongs to the concept, which is essential information for RE. Meanwhile, the concise structure of ProBase makes it convenient to be coupled with supervised relation extraction datasets. Therefore, we choose to use ProBase as our external uncertain KG and retrieve highly-related concepts of each entity. The relational facts and their confidence scores provide a prior probability distribution of concepts for an entity, which can serve as valuable supplementary knowledge given limited local contextual information.

\begin{figure}[h]
    \centering
    \includegraphics[width=1.\linewidth]{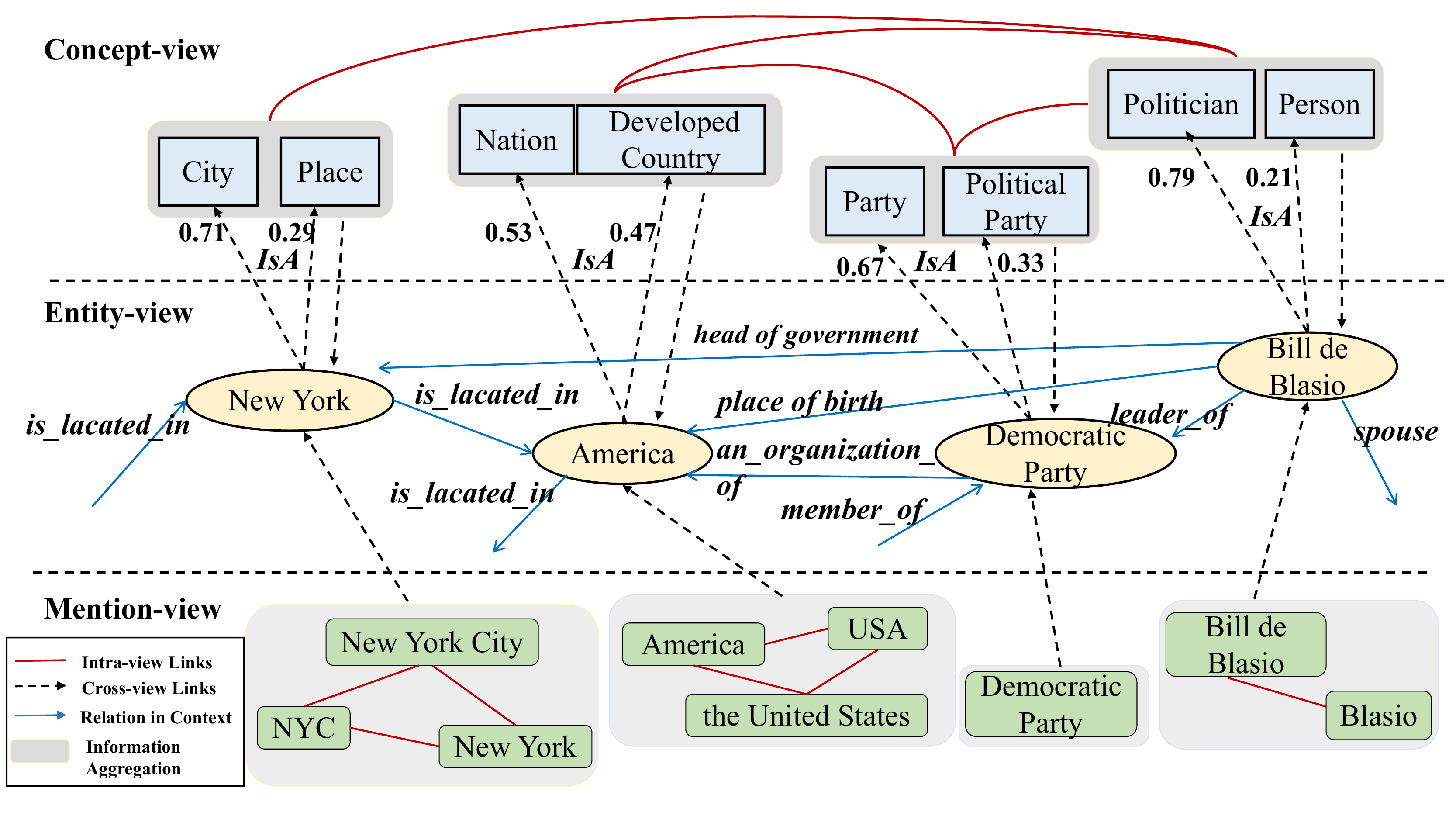}
    \caption{An example of a three-view inference framework for document-level relation extraction. In a document, an entity may have many mentions, usually in different forms. For a given entity, the local interactive representation is aggregated by different mention representations. This is a cross-view interaction between mention- and entity-view. Moreover, by retrieving uncertain KG, the concepts and weighting scores are used to produce the concept representation, which will gather descriptions of entities and concepts to obtain global interactive representations. This is a cross-view interaction between entity- and concept-view. Finally, entity-view is used for contextual information aggregation and relation prediction. Sentence-level relation extraction only involves entity- and concept-view. The figure is better viewed in color.}
    \label{intro}
\end{figure}

ProBase brings global knowledge about related concepts for an entity pair, while the target document (or sentence) provides local semantic information about related mentions. We now have three views for the contextual information of a relational fact: concept view, entity view, and mention view. It is a non-trivial task to aggregate information across different views effectively to obtain discriminative feature vectors. Inspired by \citet{DBLP:conf/kdd/HaoCYSW19}, we designed a multi-view inference framework to synthesize contextual semantics from different sources and representation levels systematically. For a given entity, we first retrieve the top $K$ concepts that the target entity most likely belongs to, and then perform cross-view interaction to aggregate the local and global information based on the confidence scores in ProBase. Figure \ref{intro} demonstrates the overview of our multi-view inference framework.

Last but not least, since the text descriptions of entities and concepts on Wikipedia also provide rich semantic information for RE, we retrieved the text descriptions from Wikipedia to enrich the representations of entities and concepts, resulting in a corpus that couples with ProBase, which we call ProBase\_Desp. The dataset provides high-quality descriptions of more than 15,000,000 entities and concepts, serving as another valuable external knowledge resource for our method. We have released ProBase\_Desp \footnote{https://github.com/pkuserc/AAAI2021-MIUK-Relation-Extraction} to facilitate further research.


In summary, we leverage uncertain KG, ProBase, to improve relation extraction for the first time. To incorporate ProBase, we design a multi-view inference mechanism that integrates local context and global knowledge across mention-, entity-, and concept-view. Experiment results show that our method achieves competitive results on both sentence- and document-level relation extraction tasks. Our contributions are summarized as follows:

\begin{itemize}
    \item We propose \textbf{MIUK}, a \textbf{M}ulti-view \textbf{I}nference framework for relation extraction with \textbf{U}ncertain \textbf{K}nowledge. Our work is pioneering research on introducing uncertain KG into relation extraction and investigating interactions among mentions, entities, and concepts.
    \item We conduct extensive experiments to verify MIUK and achieve competitive results on both sentence- and document-level relation extraction tasks. 
   \item We build and release a corpus with high-quality descriptions of more than 15,000,000 entities and concepts. It can serve as a valuable external knowledge resource for relation extraction and many other natural language processing tasks.
\end{itemize}



%% file: 3-input.tex
\section{Problem Definition and Input Formalization}

\subsection{Problem Definition}
MIUK can handle both sentence- and document-level relation extraction tasks by leveraging the multi-view inference framework. For sentence-level relation extraction, MIUK uses two-view inference framework that exploits the entity- and concept-view representations; for document-level relation extraction, the mention-view representation is added to constitute a three-view inference framework. In this section, we introduce the architecture of MIUK by taking document-level relation extraction as an example, as sentence-level relation extraction is just its simplified case. 

For an input document $D$ that contains $n$ sentences ($D = \{s_1, s_2, ..., s_n\}$), and $p$ different entities, there will be $p \cdot (p-1)$ entity pair candidates. The goal of MIUK is to predict the relations of all possible entity pairs in a parallel way. We use $m, e, c, s$ to denote the mention, entity, concept, and sentence respectively, and their corresponding low-dimensional vectors are denoted as \textbf{m, e, c, s}. Besides, the weighting score between an entity and a concept is a real number denoted as $w$.

\subsection{Input Formalization}
\begin{figure}[h]
    \centering
    \includegraphics[width=1.\linewidth]{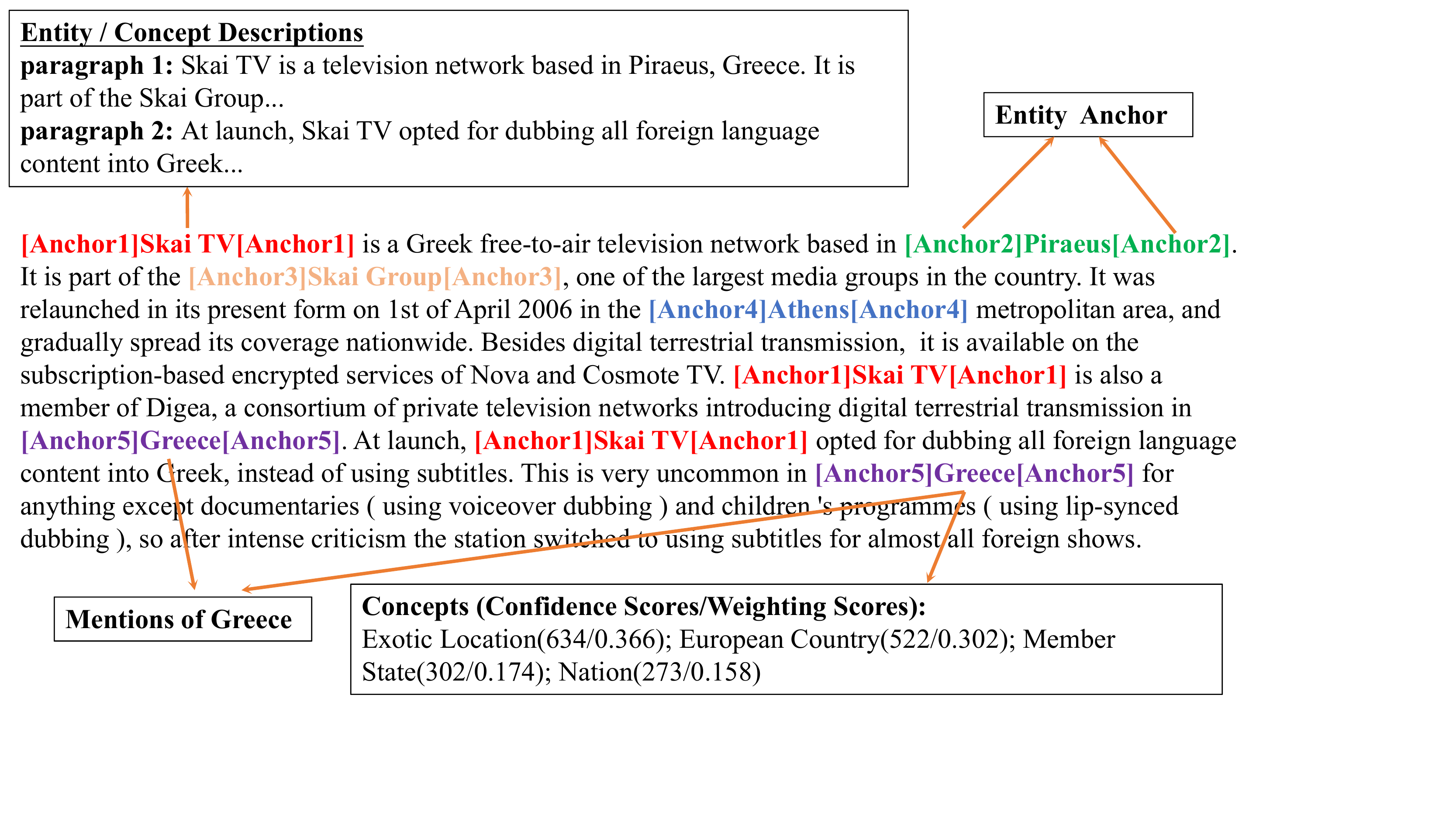}
    \caption{An illustration of entity mention, entity anchor, text descriptions, related concepts, confidence scores, and weighting scores. All the entity mentions are in bold. Different entity mentions are in different colors. There are five different entities in this document, thus 20 potential entity pairs to be predicted.}
    \label{input}
\end{figure}

This section details the data preparation for our method, which consists of three parts: input document (the context), uncertain KG (ProBase), and descriptions from Wikipedia (ProBase\_Desp), as shown in Figure \ref{input}. Note that we use Uncased BERT-Base \cite{DBLP:conf/naacl/DevlinCLT19} to encode the input document and descriptions.  

\subsubsection{Input Document} Since the input document may contain a lot of entities with various mentions, it is not applicable to add position embeddings \cite{DBLP:conf/coling/ZengLLZZ14} directly, especially when using BERT as an encoder \cite{DBLP:conf/cikm/WuH19a}. However, the position feature is crucial for relation extraction since it can highlight the target entities. In this paper, we propose to use entity anchors that can mark all the mentions and distinguish different mentions from different entities. Specifically, different entities are marked by different special tokens, while the same token indicates the various mentions of the same entity, as shown in Figure \ref{input}. These entity anchors make our model pay more attention to the entities. Each word and entity anchor is transformed into a low-dimensional word vectors by BERT.
    
\subsubsection{Uncertain KG} For each entity in the document, we retrieve the top-K concepts and their corresponding confidence scores from ProBase, which contain the prior uncertain knowledge. If an entity has less than K concepts, the token $<UNK>$ is used to pad the concept list to the fixed length, with its confidence score set to 0. 
    
\subsubsection{Entity and Concept Descriptions} For each entity and concept in ProBase, MIUK uses the first two paragraphs from Wikipedia as supplementary descriptions. If the target entity does not exist in ProBase, we then use its entity type instead. Each description is transformed into a low-dimensional vector using BERT and max-pooling operation.

%% file: 4-methods.tex
\section{Multi-view Inference}
\begin{figure*}[h]
    \centering
    \includegraphics[width=.9\linewidth]{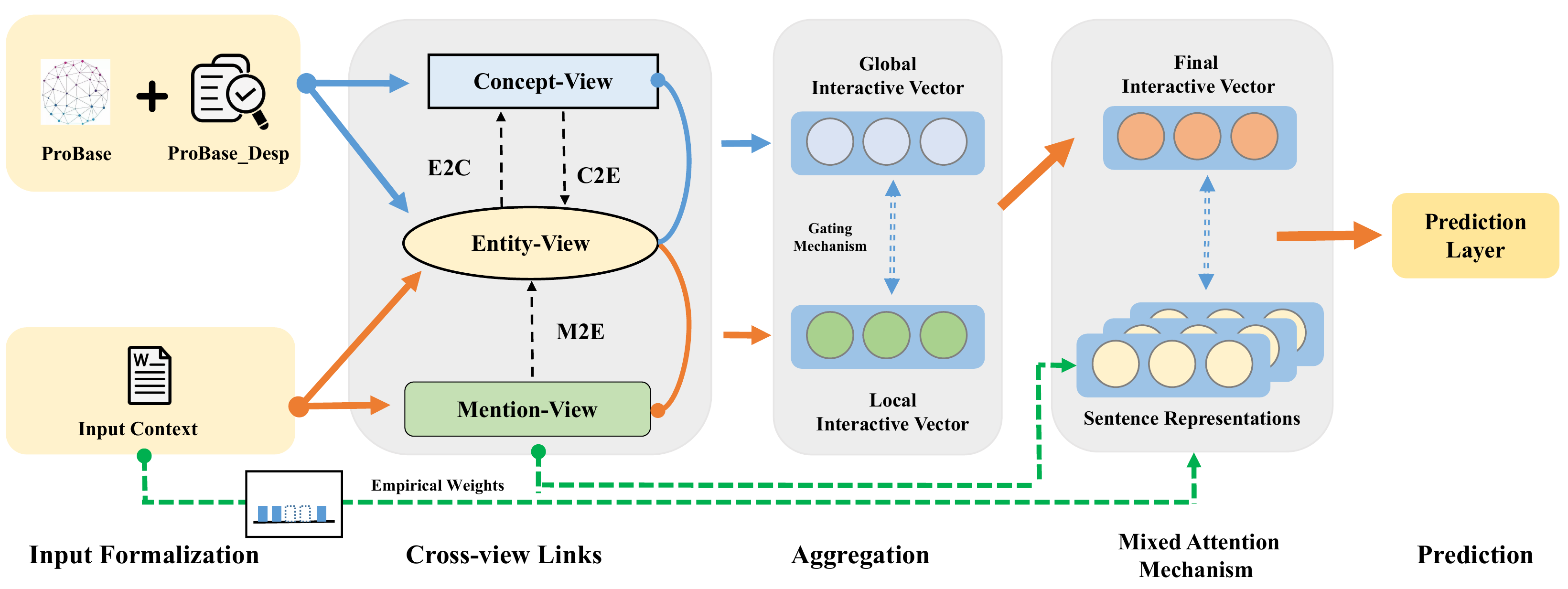}
    \caption{The neural architecture of MIUK.}
    \label{model}
\end{figure*}

The overall architecture of MIUK is shown in Figure \ref{model}. The multi-view inference framework consists of two parts: 1) cross-view links for information extraction, and 2) information aggregation and mixed attention mechanism, which aggregates the various feature vectors generated from cross-view links.

\subsection{Cross-view Links}

Cross-view links aim to extract information from the raw text effectively to generate local and global interactive vectors of entities as well as sentence representations, including the contextual information expressed by the input context and the external knowledge provided by ProBase and ProBase\_Desp.

\subsubsection{Mention2Entity Links}
Mention2Entity links (M2E) use the mention embeddings from the input document and the entity description vectors from ProBase\_Desp to obtain the local interactive vector $\textbf{u}_l$. For an entity mention ranging from the $a$-th to the $b$-th word, the mention embedding \textbf{m} is calculated by averaging the embeddings of its anchor token and all the words existing in the mention, where $\textbf{m} \in \mathbb{R}^{1 \times d}$. For a given entity $e$ with $t$ mentions $m_1, m_2, .., m_t$ in the document, unlike previous works that simply use the average of all related mention embeddings as entity representation $\textbf{e}$, we believe it is best to select the most informative mention embeddings based on the entity description from external knowledge. Therefore, MIUK employs an attention mechanism to generate the local entity representation $\textbf{e}_l$:
\begin{equation}
    \textbf{e}_l = \sum_{i=1}^{t}\alpha_{i} \cdot \textbf{m}_i;
    \alpha_{i} = \frac{exp(\textbf{e}_{d}\textbf{m}_i^T )}{\sum_{i=1}^{t}exp(\textbf{e}_{d}\textbf{m}_i^T)}.
\end{equation}
where the target entity description vector $\textbf{e}_{d} \in \mathbb{R}^{1 \times d}$ is the query vector, and $\textbf{m}_1, \textbf{m}_2, ..., \textbf{m}_t$ are the key vectors.

M2E generates the local representations of the target two entities, denoted as $\textbf{h}_l$ and $\textbf{t}_l$. We further add the minimum distance between the two target entities, denoted as $d_{ht}$ and $d_{th}$, where $d_{ht} = -d_{th}$. They are transformed into low-dimensional vectors $\textbf{d}_{ht}$ and $\textbf{d}_{th}$ by a lookup table. Finally, the local interactive vector $\textbf{u}_l$ is defined as follows:
\begin{equation}
    \textbf{u}_l = f_l([\textbf{h}_l; \textbf{d}_{ht}],[\textbf{t}_l; \textbf{d}_{th}]),
\end{equation}
where $f_l$ is the bilinear function, $[\cdot, \cdot]$ denotes concatenation operation, and $\textbf{u}_l \in \mathbb{R}^{1 \times d}$.

\subsubsection{Entity2Concept Links}
Entity2Concept links (E2C) aim to leverage the uncertain knowledge from ProBase to generate a concept vector for the target entity. 

For the target entity $e$, MIUK first retrieves the top $K$ concepts $c_1, c_2, ..., c_k$ and their corresponding confidence scores from ProBase. Then softmax operation is applied to transform the confidence scores into weighting scores $w_1, w_2, ..., w_k$, where $\sum_{i=1}^{k}{w_i} = 1$, as the confidence scores provided by ProBase are frequency counts. The corresponding concept representations are generated after encoding their descriptions and max-pooling operation, denoted as $\textbf{c}_1, \textbf{c}_2, ..., \textbf{c}_k$, where $\textbf{c}_i \in \mathbb{R}^{1 \times d}$.

Here we propose three different techniques to compute the concept vector $\textbf{c}$: 

\begin{itemize}
    \item Non-weighting Integration (NWI). NWI is a simple technique that averages all the concept representations directly. The intuition behind NWI is that all the related concepts contribute equally to target entity; thus this approach does not need the uncertain knowledge. NWI is defined as follows:
    \begin{equation}
        \textbf{c} = \frac{1}{k} \cdot \sum_{i=1}^{k}\textbf{c}_i.
    \end{equation}{}

    \item Attention-based Weighting Integration (AWI). AWI generates the concept vector $\textbf{c}$ by employing an attention mechanism. The local entity representation $\textbf{e}_l$ is the query vector, and $\textbf{c}_1, \textbf{c}_2, ..., \textbf{c}_k$ are the key vectors. AWI assumes that the local entity representation from the input document is helpful to selecting the most important concept. AWI is defined as follows:
    \begin{equation}
        \textbf{c} = \sum_{i=1}^{k}\alpha_{i} \cdot \textbf{c}_i;
        \alpha_{i} = \frac{exp(\textbf{e}_{l}\textbf{c}_i^T )}{\sum_{i=1}^{k}exp(\textbf{e}_{l}\textbf{c}_i^T)}.
    \end{equation}{}
    
    \item Prior Knowledge-based Weighting Integration (PWI). PWI uses weighting scores $w_1, w_2, ..., w_k$ to aggregate the concept representations. PWI assumes that the weighting scores provide a prior probability distribution of concepts, which is beneficial to the model's performance. PWI is defined as follows:
    \begin{equation}
        \textbf{c} = \sum_{i=1}^{k}w_i\textbf{c}_i.
    \end{equation}{}
\end{itemize}

These three techniques are based on different assumptions. We will explore their effects later in Section \ref{sec:discussion}.

\subsubsection{Concept2Entity Links}
Concept2Entity links (C2E) generate the global interactive vector $\textbf{u}_g$ by aggregating the entity description vector $\textbf{e}_d$ and the concept vector $\textbf{c}$. Specifically, for the target entities $h$ and $t$, their corresponding description vectors are denoted as $\textbf{h}_d$ and $\textbf{t}_d$. The concept vectors can be obtained from Entity2Concept links, written as $\textbf{c}_h$ and $\textbf{c}_t$. Then we use another bilinear function $f_g$ to compute the global interactive vector $\textbf{u}_g$:
\begin{equation}
    \textbf{u}_g = f_g([\textbf{h}_d; \textbf{c}_h],[\textbf{t}_d; \textbf{c}_t]),
\end{equation}
where $\textbf{u}_g \in \mathbb{R}^{1 \times d}$.

In addition, cross-view links also output all sentence representations $\textbf{s}_1, \textbf{s}_2, ..., \textbf{s}_n$ from the input document. Given a sentence ranging from the $a$-th to the $b$-th word, max-pooling operation is used to generate the sentence representation $\textbf{s}$, where $\textbf{s} \in \mathbb{R}^{1 \times d}$, which will be used in the next stage. 


\subsection{Information Aggregation and Mixed Attention Mechanism}
\subsubsection{Information Aggregation}The key insight of information aggregation is that the interactive vectors obtained from different sources can learn complementary information for the same entity pair, where $\textbf{u}_l$ contains contextual information and $\textbf{u}_g$ contains external knowledge. It is conceivable that aggregating them can achieve optimal performance. MIUK uses a gating mechanism to control the information flow between $\textbf{u}_l$ and $\textbf{u}_g$:

\begin{equation}
    \textbf{u} = \textbf{g} \odot \textbf{u}_l + (\textbf{E} - \textbf{g}) \odot \textbf{u}_g,
\end{equation}
where $\odot$ is the element-wise product between two vectors, and $\textbf{g} \in \mathbb{R}^{1 \times d}$ is the gating vector. $\textbf{E} \in \mathbb{R}^{1 \times d}$, and all the elements in $\textbf{E}$ are 1. MIUK can select the most important information and generates the final interactive vector $\textbf{u}$.

\subsubsection{Mixed Attention Mechanism} We use a mixed attention mechanism to generate the document representation $\textbf{v}$ from sentence representations $\textbf{s}_1, \textbf{s}_2, ..., \textbf{s}_n$. Intuitively, if a target entity mention exists in sentence $s$, it should be more important than other sentences. Thus we introduce an empirical weight $\gamma$ for each sentence. For the input document, mixed attention mechanism computes the weight of each sentence by combining both the information-based weight and the empirical weight:

\begin{equation}
    \textbf{v} = \sum_{i=1}^{n}(\frac{\alpha_{i} + \beta_{i}}{2} + \gamma_{i})\textbf{s}_{i},
\end{equation}

\begin{equation}
        \alpha_{i} = \frac{exp(\textbf{u}_{l}\textbf{s}_i^T )}{\sum_{i=1}^{n}exp(\textbf{u}_{l}\textbf{s}_i^T)};
        \beta_{i} = \frac{exp(\textbf{u}_{g}\textbf{s}_i^T )}{\sum_{i=1}^{n}exp(\textbf{u}_{g}\textbf{s}_i^T)},
\end{equation}{}

\begin{equation}
    \gamma_{i}=\left\{
    \begin{aligned}
    \frac{1}{z}, s_i \in S, \\
    0, s_i \notin S,
    \end{aligned}
    \right.
\end{equation}
where $\alpha_{i}$ and $\beta_{i}$ are the weights based on the local and global interactive vector; $S \in D$ and consists of $z$ different sentences, each of which contains at least one target entity mention. 

\subsection{Prediction}

Since the document-level relation extraction is a multi-label problem, i.e., a target entity pair may express more than one relation, we use a single hidden layer and sigmoid activation function for relation extraction:

\begin{equation}
    p(r|h, t) = g(\textbf{W}_r[\textbf{u}; \textbf{v}]^T + \textbf{b}_r),
\end{equation}
where $g$ is the sigmoid activation function, $\textbf{W}_r \in \mathbb{R}^{l \times 2d}$ and $\textbf{b}_r \in \mathbb{R}^{l \times 1}$ are trainable parameters, and $l$ is the number of predefined relation types in the dataset. 

%% file: 5-experiments.tex
\section{Experimental Results} 

\subsection{Datasets and Hyper-parameter Settings}
\subsubsection{Datasets and Evaluation Metrics}
For document-level relation extraction, we use DocRED proposed by \citet{DBLP:conf/acl/YaoYLHLLLHZS19}. DocRED has 3,053 training documents, 1,000 development documents and 1,000 test documents, with 97 relation types (including ``No Relation''). We treat this task as a multi-label classification problem since one or more relation types may be assigned to an entity pair. Following previous works \cite{DBLP:conf/acl/YaoYLHLLLHZS19}, we use \textbf{F1} and \textbf{IgnF1} as the evaluation metrics. IgnF1 is a stricter evaluation metric that is calculated after removing the entity pairs that have appeared in the training set.

For sentence-level relation extraction, we use ACE2005 dataset following \citet{DBLP:conf/acl/YeLXSCZ19}. The dataset contains 71,895 total instances with 7 predefined relation types (including ``No Relation''), 64,989 of which are ``No Relation'' instances (negative instances). We use five-fold cross-validation to evaluate the performance, and we report the precision (\textbf{P}), recall (\textbf{R}) and Micro F1-score (\textbf{Micro-F1}) of the positive instances.

\subsubsection{Hyper-parameter Settings}
We use uncased BERT-base \cite{DBLP:conf/naacl/DevlinCLT19} to encode the input context and text descriptions, and the size of each word embedding is 768. Note that we do not need to re-train BERT model, the rare words existing in the vocabulary of BERT can be used as the entity anchors. We then use a single layer to project each word embedding into a low-dimensional input vector of size $d$. Max-pooling operation is further applied to compute the entity description vector or the concept representation. Note that we use the same BERT model to encode both the input context and the entity/concept descriptions. We experiment with the following values of hyper-parameters: 1) the learning rate $lr_{BERT}$ and $lr_{Other}$ for BERT and other parameters $\in \{1\times10^{-3}, 1\times10^{-4}, 1\times10^{-5}\}$; 2) the size of input vector, entity description vector and concept representation $\in \{50, 100, 150, 200\}$; 3) the size of distance embedding $\in \{5, 10, 20, 30\}$; 4) batch size $\in \{4, 8, 12, 16, 20, 24\}$; and 5) dropout ratio $\in \{0.1, 0.2, 0.3, 0.4, 0.5\}$. We tune the hyper-parameters on the development set, and we evaluate the performance on the test set. Table \ref{para} lists the selected hyper-parameter values in our experiments.

\begin{table}[h]
	\renewcommand
	\renewcommand{\multirowsetup}{\centering}
	\centering
	\setlength{\tabcolsep}{3mm}\begin{tabular}{|c|c|}
		\hline
		\textbf{Hyper-parameter}&\textbf{Value}\\\hline
		$lr_{BERT}$&$1\times10^{-5}$\\
		$lr_{Other}$&$1\times10^{-5}$\\
		Input Vector Size $d$ &100\\
		Embedding Size of Entity/Concept &100\\
		Embedding Size of Distance &10\\
		Batch Size &16\\
		Dropout Ratio &0.2
		\\\hline
	\end{tabular}
	\caption{Hyper-parameter Settings}
	\label{para}
\end{table}

\subsection{Baseline Models}
We choose a number of methods as baseline models for sentence- and document-level relation extraction.

\subsubsection{Document-level Relation Extraction} The following models are used as baseline models. \textbf{GCNN} \cite{DBLP:conf/acl/SahuCMA19} and  \textbf{EoG} \cite{DBLP:conf/emnlp/ChristopoulouMA19} are graph-based document-level relation extraction models. In addition, the following four methods all use BERT-base as encoder: \textbf{BERT-Two-Step}  \cite{DBLP:journals/corr/abs-1909-11898} classified the relation between two target entities in two steps; \textbf{DEMMT} \cite{DBLP:journals/access/HanW20} proposed an entity mask method, which is similar to entity anchor; \textbf{GEDA} \cite{DBLP:conf/coling/LiYSXXZ20} proposed a graph-enhanced dual attention network for document-level
relation extraction; \textbf{LSR} \cite{DBLP:conf/acl/NanGSL20} used a multi-hop reasoning framework with external NLP tools, which is the state-of-the-art model for document-level relation extraction.
    
\subsubsection{Sentence-level Relation Extraction} The following four approaches are used as baseline models: \textbf{SPTree} \cite{DBLP:conf/acl/MiwaB16} used tree-LSTM for relation extraction; \textbf{Walk-based Model} \cite{DBLP:journals/corr/abs-1902-07023} proposed a graph-based model that considers interactions among various entities; \textbf{MLT\_Tag} \cite{DBLP:conf/acl/YeLXSCZ19} exploited entity BIO tag embeddings and multi-task learning for sentence-level relation extraction. Furthermore, we use BERT to replace the encoder in MLT\_Tag to form an additional strong baseline named \textbf{BERT-MLT\_Tag}.


\subsection{Main Results}
\subsubsection{Document-level Relation Extraction}

Table \ref{doc} records the performance of the proposed MIUK and other baseline models. We can see that:

\begin{table}[h]
    \centering
    \setlength{\tabcolsep}{1.5mm}\begin{tabular}{ccc|cc}
    \midrule
    \multirow{2}{*}{\textbf{Methods}} &\multicolumn{2}{c}{\textbf{Dev}} & \multicolumn{2}{c}{\textbf{Test}} \\
    \cmidrule(r){2-3} \cmidrule(r){4-5}
    &\textbf{IgnF1\%}&\textbf{F1\%}&\textbf{IgnF1\%}&\textbf{F1\%}  \\
    \midrule
    \textbf{GCNN}$\S$&46.22&51.52&49.59&51.62\\
    \textbf{EoG}$\S$&45.94&52.15&49.48&51.82\\
    \textbf{BERT-Two-Step}&-&54.42&-&53.92\\
    \textbf{GEDA}&54.52&56.16&53.71&55.74\\
    \textbf{DEMMT}&55.50&57.38&54.93&57.13\\
    \textbf{LSR}$\S$&52.43&59.00&56.97&59.05\\
    \textbf{MIUK (three-view)}&\textbf{58.27}&\textbf{60.11}&\textbf{58.05}&\textbf{59.99}\\
    \bottomrule
    \end{tabular}
    \caption{Document-level relation extraction results on the development set and the test set of DocRED. MIUK (three-view) is designed for document-level relation extraction that contains mention-, entity- and concept-view representations. Results with $\S$ are directly cited from \cite{DBLP:conf/acl/NanGSL20}.}
    \label{doc}
\end{table}

\begin{itemize}
    \item The proposed MIUK (three-view) outperforms other models by a large margin in terms of both F1 and IgnF1. Specifically, compared with the highest scores among baseline models, MIUK achieves 1.11 and 1.08 absolute increase in F1 score on both the development and the test set. Similarly, the absolute increases in IgnF1 score achieved by MIUK are 2.77 and 0.94. The results can verify the effectiveness of our multi-view inference architecture with uncertain knowledge incorporated.
    
    \item MIUK and LSR outperform other models (e.g., BERT-Two-Step, GEDA, and DEMMT) by a large margin, even though the latter involve novel training skills or sophisticated architectures (e.g., graph neural networks). We mainly attribute this to the introduction of external knowledge: LSR brings in syntactic dependence information based on external tools, while MIUK incorporates rich information from ProBase and Wikipedia.
    
    \item Though LSR achieves impressive results with NLP tools on the test set, the IgnF1 score of LSR on the development set is merely 52.43, far lower than its IgnF1 score on the test set. One possible explanation for this instability could be the usage of external NLP tools (LSR used spaCy\footnote{https://spacy.io/} to get meta dependency paths of sentences in a document), as external tools may cause error propagation problems. Our method achieves satisfactory results on both the development and the test set simultaneously, showing that leveraging uncertain knowledge can not only boost the model's performance but also improve its generalization ability.
\end{itemize}

\begin{table}[h]
	\renewcommand
	\renewcommand{\multirowsetup}{\centering}
	\centering
	\setlength{\tabcolsep}{3mm}\begin{tabular}{c|c|c|c}
		\hline
		\textbf{Methods}&\textbf{P\%}&\textbf{R\%}&\textbf{Micro-F1\%}
		\\\hline
		\textbf{SPTree}\dag&70.1&61.2&65.3\\
		\textbf{Walk-based Model}\dag&69.7&59.5&64.2\\
		\textbf{MTL\_Tag}\dag&66.5&71.8&68.9\\
		\textbf{BERT-MTL\_Tag}*&70.1&74.5&72.0\\
		\textbf{MIUK (two-view)}&74.7&76.9&75.7\\
		\hline
	\end{tabular}
	\caption{Comparison between MIUK and the state-of-the-art methods using ACE 2005 dataset for sentence-level relation extraction. The best results are in bold. MIUK (two-view) is used for sentence-level relation extraction that consists of entity- and concept-view representations. Models with $*$ are reproduced based on open implementation. Results with $\dag$ are directly cited from \cite{DBLP:conf/acl/YeLXSCZ19}.}
	\label{sent}
\end{table}

\subsubsection{Sentence-level Relation Extraction}

For sentence-level relation extraction, MIUK (two-view) outperforms other models by a large margin with an F1 score of 75.7, which clearly sets up a new state-of-the-art. Besides, MIUK also achieves the best \textbf{P} (74.7) and \textbf{R} (76.9).  Compared with BERT-MTL\_Tag, a competitive model equipped with BERT, our method still achieves higher F1 score by 3.7, \textbf{P} by 4.6, and \textbf{R} by 2.4 in absolute value, while balancing \textbf{P} and \textbf{R} better. These results show that the introduction of external knowledge and our multi-view inference framework that utilizes uncertain knowledge can also benefit sentence-level relation extraction.


\subsection{Detailed Analyses}\label{sec:discussion}

Our model mainly involves leveraging uncertain knowledge, multi-view inference, and text descriptions. We will further evaluate the effectiveness of these three components in this section. Due to space limit, we only report the detailed analyses on document-level relation extraction; similar conclusions can be drawn from sentence-level relation extraction as well.


\begin{table}[h]
	\renewcommand
	\renewcommand{\multirowsetup}{\centering}
	\centering
	\setlength{\tabcolsep}{7mm}\begin{tabular}{c|c}
		\hline
		\textbf{Models}&\textbf{F1\%}\\
		\hline
		\textbf{MIUK}&\textbf{60.11}\\
		\hline
		\textbf{MIUK-NWI}&59.37\\
		\textbf{MIUK-AWI}&59.42\\
		\hline
		\textbf{MIUK w/o Cross-view Inference}&58.03\\
		\textbf{MIUK w/o Mixed Att}&59.70\\
		\hline
		\textbf{MIUK w/o Entity Desp}&58.21\\
		\textbf{MIUK w/o Concept Desp}&59.02\\
		\hline
	\end{tabular}
	\caption{The F1 scores of MIUK and its variants on the development set of DocRED.}
	\label{ablation}
\end{table}

\begin{table}[h]
	\renewcommand
	\renewcommand{\multirowsetup}{\centering}
	\centering
	\setlength{\tabcolsep}{6mm}\begin{tabular}{c|c|c}
		\hline
		\textbf{Top-K}&\textbf{IgnF1\%}&\textbf{F1\%}\\
		\hline
		\textbf{K = 1}&55.65&58.22\\
		\textbf{K = 2}&57.03&59.35\\
		\textbf{K = 3}&\textbf{58.27}&\textbf{60.11}\\
		\textbf{K = 4}&57.22&59.21\\
		\textbf{K = 5}&55.40&58.53\\
		\hline
	\end{tabular}
	\caption{The results of using different top $K$ concepts for each entity in DocRED.}
	\label{k}
\end{table}
\subsubsection{Effectiveness of Uncertain Knowledge} The main difference between deterministic KGs and uncertain KGs is whether relations of entity pairs are assigned with confidence scores. To explore the effectiveness of uncertain knowledge, we design a model variant named \textbf{MIUK-NWI} that uses deterministic knowledge. The only difference between MIUK-NWI and MIUK is that MIUK-NWI uses a KG with all confidence scores removed (or set as the same). As shown in Table \ref{ablation}, we can observe a notable drop in performance (nearly 1.0 in F1 score) comparing MIUK-NWI with MIUK, which shows that the prior probability distribution of concepts for an entity provides valuable global contextual information, and our framework is capable of capturing this information to discriminate relational facts. 

We further design \textbf{MIUK-AWI} as an enhanced version of MIUK-NWI, which uses a classic attention mechanism to aggregate concept information in a deterministic KG. We find the the performances of MIUK-AWI and MIUK-NWI show no clear difference. We can roughly conclude that compared with the classic attention mechanism, the prior confidence scores can help identify relevant concepts better.

Though ProBase provides a number of concepts for a given entity, using too many concepts may also bring noisy information and thus hinder the model's performance. Therefore, the concept number $K$ for our MIUK is also an important hyper-parameter worth investigating. Table \ref{k} shows the results using different $K$ for concept selection. As expected, using too many concepts (more than three) does not gain better results. If only one or two concepts are selected, we can also observe performance degradation, due to feeding limited external knowledge into the inference framework. In short, we find that $K = 3$ generates the best results in our experiments.   
    
\subsubsection{Effectiveness of Multi-view Inference} To verify the effectiveness of the multi-view inference framework that we design, we build a model variant named \textbf{MIUK w/o Cross-view Inference}, which simply concatenate all the representations from different sources and then feed them into the final classifier. From Table \ref{ablation} we can see that removing the multi-view inference framework results in significant performance degradation (more than 2 in F1 score). The result shows that our multi-view inference framework provides a better way to integrate and synthesize multi-level information from different sources.

The mixed attention mechanism is another important component of MIUK. We further design a model variant named \textbf{MIUK w/o Mixed Att}, which replaces the mixed attention mechanism with the vanilla attention mechanism. Table \ref{ablation} shows that the mixed attention mechanism improves the F1 score by 0.41. One possible explanation is that when the input document contains too many sentences, the vanilla attention mechanism fail to focus on highly-related sentences. We conclude that the mixed attention mechanism with empirical weights can capture supplementary information that is independent of contextual information.

\subsubsection{Effectiveness of Text Descriptions} To investigate how the information in text descriptions affects our model, we create two variants of MIUK by removing entity descriptions or concept descriptions, named \textbf{MIUK-w/o Entity Desp} and \textbf{MIUK-w/o Concept Desp}, respectively. From Table \ref{ablation} we can see that the F1 scores drop significantly without either entity descriptions (MIUK-w/o Entity Desp) or concept descriptions (MIUK-w/o Concept Desp), which shows that the information from Wikipedia benefits relation extraction and MIUK can capture the rich semantics in these text descriptions well.

%% file: 2-related-work.tex
\section{Related Work}

\subsection{Uncertain Knowledge Graphs}
Uncertain Knowledge Graphs provide a confidence score for each word pair, such as ConceptNet and ProBase. ConceptNet is a multilingual uncertain KG for commonsense knowledge. It gives a list of words with certain relations (such as ``located at,'' ``used for,'' etc.) to the given entity, and provides a confidence score for each word pair. ProBase, otherwise known as Microsoft Concept Graph, is a big uncertain KG with 5,401,933 unique concepts, 12,551,613 unique entities and 87,603,947 \textit{IsA} relations. For a given entity and concept pair that has \textit{IsA} relation, denoted as $(x, y, P_{IsA}(x,y))$, $P_{IsA}(x,y)$ is the confidence score that measures the possibility of the entity $x$ belonging to the concept $y$. 

Since ProBase provides a more concise data structure and is easier to apply to relation extraction, we choose ProBase as the source of uncertain knowledge in this paper.

\subsection{Relation Extraction with External Knowledge}
Most external knowledge-based methods are targeted at distant supervision relation extraction (DSRE). \citet{DBLP:conf/akbc/VergaM16} employed FreeBase \cite{DBLP:conf/sigmod/BollackerEPST08} and probabilistic model to extract features for DSRE. \citet{DBLP:conf/emnlp/WestonBYU13} first proposed to use both contextual and knowledge representations for DSRE, but they used the two representations independently, and connected them only at the inference stage. \citet{DBLP:conf/aaai/Han0S18}, and \citet{DBLP:conf/naacl/XuB19} designed heterogeneous representation methods which jointly learn contextual and knowledge representations. \citet{DBLP:conf/coling/LeiCLDY0S18} proposed a novel bi-directional knowledge distillation mechanism with a dynamic ensemble strategy (CORD). For each entity, CORD uses the related words from FreeBase by n-gram text matching, which may bring lots of noise. 

Some works leverage external knowledge for supervised relation extraction. \citet{DBLP:conf/coling/RenZLLZLL18} used the descriptions of entities from Wikipedia but did not incorporate KGs for further research. \citet{DBLP:conf/acl/LiDLZS19} only focused on Chinese relation extraction with HowNet \cite{dong2003hownet}.  \citet{DBLP:conf/emnlp/LiMYL19} incorporated prior knowledge from external lexical resources into a deep neural network. For each relation type, they found all relevant semantic frames from FrameNet and their synonyms from Thesaurus.com. However, they only considered the keywords and synonyms of an entity; therefore, the rich information in the entity description was ignored. MIUK distinguishes itself from previous works by introducing ProBase into relation extraction and systematically investigating the interactions among mentions, entities, and concepts.

